\documentclass[10pt]{article}
\usepackage[margin=24mm]{geometry}
\usepackage[T1]{fontenc}
\usepackage{lmodern,microtype,amsmath,amssymb,booktabs,array,xcolor}
\usepackage{tikz,graphicx}
\usetikzlibrary{arrows.meta,positioning,fit}
\usepackage[colorlinks=true,linkcolor=blue!55!black,citecolor=blue!55!black,urlcolor=blue!55!black]{hyperref}
\usepackage{enumitem,caption}
\setlist{nosep,leftmargin=*}
\definecolor{jevblue}{HTML}{0072B2}
\definecolor{denseorange}{HTML}{D55E00}
\definecolor{hybridgreen}{HTML}{009E73}
\definecolor{ink}{HTML}{263445}
\definecolor{papergray}{HTML}{F3F5F7}
\color{ink}
\newcommand{\method}[1]{\texttt{#1}}
\definecolor{studybm25}{HTML}{000000}
\definecolor{studyopenaidense}{HTML}{D55E00}
\definecolor{studybgem3}{HTML}{CC79A7}
\definecolor{studyhybrid}{HTML}{E69F00}
\definecolor{studyhybridbge}{HTML}{009E73}
\definecolor{studyhybridjev}{HTML}{56B4E9}
\definecolor{studyjevscan}{HTML}{0072B2}
\definecolor{decisionfigopenaidense}{HTML}{D55E00}
\definecolor{decisionfighybridbge}{HTML}{009E73}
\definecolor{decisionfighybridjev}{HTML}{56B4E9}
\definecolor{decisionfigjevscan}{HTML}{0072B2}
\definecolor{decisionfighybriddecisions}{HTML}{AA4499}
\definecolor{decisionfigdecisionsscan}{HTML}{882255}
\definecolor{clefhybrid}{HTML}{117733}
\definecolor{clefscan}{HTML}{44AA99}
\definecolor{clefjev}{HTML}{0072B2}
\definecolor{clefdecisions}{HTML}{AA4499}
\title{\textbf{Scan or Index?}\\[4pt]\large Evidence Coverage, Answer Quality, and Cost in Long-Document RAG}
\author{
Juan Terven\textsuperscript{1}, Diana-Margarita Córdova-Esparza\textsuperscript{2},Julio-Alejandro Romero-González\textsuperscript{2}\\
Edgar Chavez-Urbiola\textsuperscript{1}, Juan Irving Vasquez\textsuperscript{3}, Julian Andres Ramirez-Bautista\textsuperscript{4}\\
Silvia L. Chaparro-Cárdenas\textsuperscript{5}, Gendry Alfonso-Francia\textsuperscript{6}\\[6pt]
\textsuperscript{1}CICATA Qro., Instituto Politécnico Nacional, México\\
\textsuperscript{2}Facultad de Informática, Universidad Autónoma de Querétaro, México\\
\textsuperscript{3}CIDETEC, Instituto Politécnico Nacional, México\\
\textsuperscript{4}Fundación Universitaria de San Gil - Unisangil, Colombia\\
\textsuperscript{5}Universidad de Investigación y Desarrollo, Colombia\\
\textsuperscript{6}ENES, Universidad Nacional Autónoma de México
}
\date{October 9, 2026}
\begin{document}
\maketitle

\begin{abstract}
Scanning every passage in a long document seems thorough, but does it preserve more useful evidence or produce better answers? We compare 11 retrieval configurations spanning lexical and dense search, hybrid reranking, and full-document scanning with Jev, OpenAI Decisions, and Cloudflare Clef. Our evaluation uses 396 QASPER questions and 80 questions from eight complete textbooks, with common passages, a five-passage limit, and a fixed answer generator. On books, Hybrid + Jev achieves the highest observed evidence coverage (94.1\%), compared with 88.7\% for Jev full scan and 81.3\% for OpenAI dense retrieval. Full scans examine every passage initially, yet later reduction steps can discard relevant evidence. Higher evidence coverage also does not consistently produce better answers: Hybrid + Decisions achieves the highest observed book Answer Accuracy among retrieval methods (0.892), despite recovering less reference evidence. Full scanning incurs substantially higher latency on books under the tested deployments, while reusable dense retrieval has lower projected API cost than paid decision-based methods as queries accumulate. Most QASPER articles already fit within the hybrid candidate pool, limiting the benefit of scanning. These results highlight evidence retention, answer quality, and reuse cost as complementary criteria for choosing a long-document retrieval strategy.
\end{abstract}

\section{Introduction}

Answering a question about a textbook can mean finding a few useful passages among hundreds of pages. Retrieval-augmented generation (RAG) supplies such passages to a language model before it writes an answer~\cite{rag,fid}. How those passages are found affects both the available evidence and the work repeated for every question. Dense retrieval encodes passages in advance and reuses their representations across queries~\cite{dpr}. Its preparation cost is therefore spread across later searches.

Decision models offer another way to select evidence. Jev chooses among supplied options and can use passages as those options~\cite{jeveval,jevfind}. It can rerank candidates from an existing index or examine every passage for each new query. The latter avoids precomputing passage embeddings, but long documents must be processed in windows and reduced to a final selection. Evidence can disappear during this reduction even when every passage was examined initially. The practical question is whether broader access improves the final evidence enough to justify the repeated work.

We examine this tradeoff within a specified document using 396 QASPER questions with human references and 80 questions about eight complete textbooks. The original study compares seven lexical, dense, hybrid, and Jev-based methods with common passage boundaries and a five-passage retrieval limit. Separate follow-up evaluations add OpenAI Decisions~\cite{decisionsdocs} and local Cloudflare Clef~\cite{clefmodel}. Almost all sampled articles fit inside the shared 50-candidate pool, whereas every book requires substantial preselection or reduction. The two corpora, therefore, test different retrieval conditions.

Our contribution is an empirical comparison that connects retained evidence, reduction losses, and cost under document reuse. We locate reference evidence in the source, trace what survives each selection stage, and account for preparation and query API charges. A fixed answer generator provides a separate test of whether retrieval differences carry through to automated answer scores. On the book sample, Hybrid + Jev retains the most annotated evidence on average while requiring much less query time than Jev full scanning in the measured deployment. The results make candidate retention and document reuse central to the comparison, while keeping evidence coverage distinct from answer correctness.

\subsection{Related work}

Passage retrieval ranges from term matching to neural representations. BM25 provides a lexical baseline~\cite{bm25}, while Sentence-BERT, Dense Passage Retrieval, and Contriever support similarity search with independently encoded text~\cite{sbert,dpr,contriever}. ColBERT retains token-level representations for later interaction with the query~\cite{colbert}. BGE-M3 supports dense, sparse, and multi-vector retrieval~\cite{bgem3}; our experiments use its dense mode. Reciprocal rank fusion combines ranked lists without requiring their scores to share a scale~\cite{rrf}. These methods provide established ways to reuse document representations against which repeated passage scoring can be compared.

Reranking spends additional computation on a candidate set. Unsupervised Passage Re-ranking (UPR) scores passages using the likelihood of generating the question~\cite{upr}. RankGPT uses language models for listwise ranking~\cite{rankgpt}, pairwise ranking prompting compares two passages at a time~\cite{pairwise}, and ListT5 uses bounded comparisons with tournament selection~\cite{listt5}. These studies establish model-based passage comparison as a retrieval component. Our focus is how reference evidence survives when such selection is repeated across a full document.

Long-document methods also differ in what they present to the model. RAPTOR builds a hierarchy of clustered summaries~\cite{raptor}, while STAIR uses document structure to retrieve relevant sections~\cite{stair}. LongBench and RULER evaluate long-context capabilities across tasks and input lengths~\cite{longbench,ruler}, and Lost in the Middle shows that evidence position can affect its use~\cite{lostmiddle}. Comparisons of RAG and long-context generation find that quality and cost depend on the model and task~\cite{raglc,lara}. Our full scan repeatedly selects original passages and supplies only the finalists to the generator; it does not give the generator the entire document.

Prior Jev studies include a community evaluation on a six-book reconstruction of SearchTome and a Turkish XQuAD report with a full-corpus hierarchical ablation~\cite{jevbooks,jevxquad}. An academic evaluation examines Jev across classification, reasoning, and other decision tasks~\cite{jeveval}. The community reports provide practical precedents, with different datasets and protocols. We build on this work by measuring how much annotated source text remains, tracing reduction losses, and including preparation costs when estimating API charges for repeated queries.

Evaluation must also distinguish retrieval from answer generation. BEIR and MTEB demonstrate the need to evaluate retrieval and embeddings across tasks~\cite{beir,mteb}; QASPER provides human questions, answers, and supporting evidence within research papers~\cite{qasper}. The RGB benchmark examines evidence integration, noise robustness, and rejection of unsupported questions~\cite{rgb}. RAGAs and RAGBench assess complementary properties of retrieved context and generated answers~\cite{ragas,ragbench}, while ALCE evaluates whether answers are supported by citations~\cite{alce}. Our source-span metric measures recovery of annotated evidence. It can miss alternative supporting passages, and neither this metric nor our answer scores verifies citation correctness.

Reference provenance further affects what an evaluation can establish. mtRAG includes human conversations and unanswerable questions~\cite{mtrag}, while work on synthetic RAG evaluation finds that agreement with human evaluation depends on the setting~\cite{syntheticrag}. ARES uses human annotations to correct automated estimates~\cite{ares}. Model judgments can also exhibit self-preference and position bias~\cite{selfpreference,positionbias}. We therefore report articles with human references separately from textbooks with frontier model-generated questions and interpret answer scores within the limits of model-based evaluation.

\section{Experimental design}

We follow evidence from the source document to the final answer. Retrieval methods share the same documents, questions, passages, and answer generator, allowing us to compare evidence retention, latency, API cost, and answer quality. Experimental code, configurations, and aggregate results are available at \url{https://github.com/jrterven/jev-vs-rag}.

\subsection{Data and retrieval setup}

We use 396 QASPER questions from 264 research articles: 320 answerable and 76 unanswerable~\cite{qasper}. Selected questions have consistent answerability annotations. Answerable cases must include a complete set of textual evidence that can be located in the source; ambiguous matches and evidence requiring images or tables are excluded. Each answerable question keeps a fixed human reference answer for the later answer evaluation.

The book set contains 80 questions across eight complete OpenStax textbooks, with four factual, three explanatory, and three multi-section questions per book. These questions and reference answers were primarily generated with the frontier model GPT-Astra from relevant sections and neighboring text, without access to retrieval outputs or evaluation scores. Supporting quotations were checked against the source. These checks establish textual alignment, rather than independent human validation. A separate development set contains 60 QASPER training questions and 20 questions from two additional textbooks. Sources and references were fixed before testing. Rephrasing questions does not rule out training-data overlap~\cite{rephrased}.

Retrieval takes place within the specified document. We process the complete book text, including available figure captions and alternative text. All methods use the \method{cl100k\_base} tokenizer to form passages of approximately 500 tokens with 75-token overlap, and return up to five passages. Articles contain 2--70 passages (median 11), compared with 662--2,245 per book. Almost all articles therefore fit inside the shared pool of up to 50 hybrid candidates, while every book requires substantial preselection or reduction.

\begin{figure}[!htbp]
\centering
\begin{tikzpicture}[
  x=1mm,y=1mm,
  font=\fontsize{8.5}{10}\selectfont,
  route/.style={draw=black!55,line width=.45pt,fill=black!2,
    text width=57mm,minimum height=11mm,inner sep=1.5mm,align=center},
  endpoint/.style={draw=black!55,line width=.45pt,fill=black!2,
    inner sep=1.5mm,align=center},
  wire/.style={draw=black!65,line width=.55pt},
  arrow/.style={wire,-{Stealth[length=1.7mm,width=1.2mm]}}
]
\node[endpoint,text width=24mm,minimum height=18mm] (source) at (13.5,0)
  {Document + question\\Common passages\\500 tokens, 75 overlap};
\node[route] (indexed) at (64,14)
  {\textbf{Indexed retrieval}\\Reusable index $\rightarrow$ rank};
\node[route] (hybrid) at (64,0)
  {\textbf{Hybrid + reranking}\\RRF $\rightarrow$ up to 50 candidates $\rightarrow$ rerank};
\node[route] (scan) at (64,-14)
  {\textbf{Full scan}\\All passages $\rightarrow$ windows $\rightarrow$ reduce\\Up to 10 candidates per window};
\node[endpoint,text width=17mm,minimum height=12mm] (selected) at (109,0)
  {Selected\\passages\\Up to five};
\node[endpoint,text width=24mm,minimum height=12mm] (generator) at (141,0)
  {Fixed answer\\generator};
\draw[wire] (source.east) -- (31,0);
\draw[wire] (31,-14) -- (31,14);
\foreach \name/\level in {indexed/14,hybrid/0,scan/-14} {
  \draw[arrow] (31,\level) -- (\name.west);
  \draw[wire] (\name.east) -- (97,\level);
}
\draw[wire] (97,-14) -- (97,14);
\draw[arrow] (97,0) -- (selected.west);
\draw[arrow] (selected.east) -- (generator.west);
\end{tikzpicture}
\caption{The three retrieval routes share a document, question, and passage boundaries. Indexed retrieval ranks from a reusable index; hybrid reranking selects from a common pool of up to 50 candidates; full scanning considers every passage before hierarchical reduction. Each route returns up to five passages, packed under the same 2,500-token budget for the fixed generator.}
\label{fig:retrieval-strategies}
\end{figure}
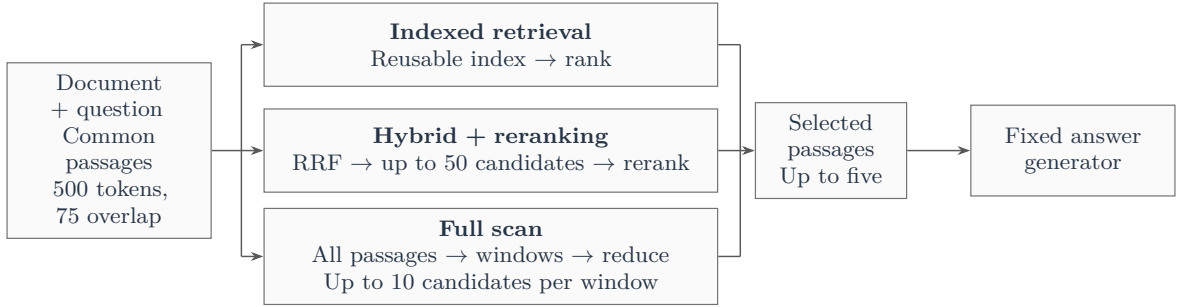

Figure~\ref{fig:retrieval-strategies} outlines the three retrieval routes; Table~\ref{tab:methods} summarizes the 11 configurations. Jev, Decisions API, and Clef are each evaluated in two settings: selecting from the shared pool of up to 50 hybrid candidates, and scanning all passages before reducing them to the final results.

\begin{table}[ht]
\centering\small
\begin{tabular}{p{.24\linewidth}p{.67\linewidth}}
\toprule
Method & Configuration \\
\midrule
\multicolumn{2}{l}{\textit{Retrieval baselines}} \\
BM25 & $k_1=1.2$, $b=0.75$; English tokenization and stemming \\
OpenAI dense & \method{text-embedding-3-small}; exact cosine search \\
BGE-M3 dense & \method{BAAI/bge-m3}; exact cosine search \\
Hybrid RRF & BM25 + BGE-M3, combined by reciprocal rank fusion (constant 60) \\
\midrule
\multicolumn{2}{l}{\textit{Hybrid reranking}} \\
Hybrid + BGE & Up to 50 hybrid candidates, reranked with \method{bge-reranker-v2-m3} \\
Hybrid + Jev & The same hybrid candidates, selected with Jev \\
Hybrid + Decisions & The same hybrid candidates, selected with Decisions API \\
Hybrid + Clef & The same hybrid candidates, selected with local Clef \\
\midrule
\multicolumn{2}{l}{\textit{Full scans}} \\
Jev full scan & Every passage initially, followed by Jev selection and reduction \\
Decisions full scan & Every passage initially, followed by Decisions selection and reduction \\
Clef full scan & Every passage initially, followed by local Clef selection and reduction \\
\bottomrule
\end{tabular}
\caption{The 11 retrieval configurations. All use common passages and return up to five results; hybrid rerankers receive the same candidate pool.}
\label{tab:methods}
\end{table}

We use Jev (\method{jev-1.13.0}), OpenAI Decisions API (\method{gpt-6-luna}), and the 27B \method{Cloudflare/clef} model, with Clef running locally~\cite{jevdocs,decisionsdocs,clefmodel}. All three process passages in windows with a 12,000-token request limit, including overhead. Jev uses conservative token estimates, Decisions counts its serialized requests, and Clef checks both native and common-token counts. Decisions also includes an explicit no-match option, which is excluded from the returned passages.

Each window keeps up to ten candidates. Their complete text is evaluated again in successive rounds until a final comparison returns up to five passages. Scores are used within each comparison and are not treated as global scores across windows~\cite{jevfind}. Clef uses the probabilities from its released implementation, with source-order tie breaking. In all three full-scan variants, every passage is considered initially, but later reductions can still discard relevant evidence.

\subsection{Evidence, latency, and API cost}

Evidence coverage measures how much annotated support appears in the retrieved passages. Let $R_k$ be the union of source-token intervals covered by the first $k$ passages, and let $S_j$ be one complete alternative set of reference evidence units:
\[
\mathrm{EC}@k=
\max_j\frac{1}{|S_j|}
\sum_{e\in S_j}\frac{|e\cap R_k|}{|e|}.
\]
Evidence units receive equal weight, overlapping tokens count once, and units from different alternatives are never combined. Complete evidence recovery requires one entire alternative set. An exact oracle gives the best attainable coverage with up to five passages. These metrics apply to answerable questions; failed assigned retrievals receive zero coverage. For Jev, Decisions, and Clef, reduction traces track evidence through the initial candidate pool, first selection, final candidate pool, and returned passages.

We average scores within each document and then across documents. Paired differences use the same questions and 10,000 bootstrap resamples of whole documents. We report descriptive pointwise 95\% intervals without adjustment for multiple comparisons.

We report median and 95th-percentile latency over valid completed executions. The comparison among Jev, Decisions, and Clef uses the first retrieval execution for each question in the full test set. Hybrid times cover reranking of the shared candidate pool, excluding candidate generation; full-scan times cover passage selection and reduction. Clef latency summaries use these single executions. End-to-end measurements for BM25, dense and hybrid baselines, Jev, and Decisions also cover a 100-question subset, equally divided between corpora, with three assigned executions per question and randomized method order. These measurements include query embeddings, candidate generation, and ranking. Jev summaries pool valid repetitions; Decisions summaries distinguish exclusive and shared GPU availability.

Document preparation and answer generation are timed separately. Measured retrieval uses no automatic retries or query-result cache. BGE models run on an NVIDIA GB10 in FP32 with batch size eight. Clef runs on a separate shared GB10 in BF16, with one forward pass at a time; Jev and Decisions use two concurrent API calls. Latencies describe these deployments and their hardware, precision, concurrency, and service conditions.

API charges use reported consumption and the prices recorded during execution. Let $P_m$ be the mean preparation cost per document and $\overline{Q_m}$ the mean cost per assigned base query for method $m$. For $N$ queries, total and average cost are projected as
\[
C_m(N)=P_m+N\overline{Q_m},
\qquad
A_m(N)=\frac{C_m(N)}{N}.
\]
Query costs are averaged within documents and then across documents, including unanswerable queries and recorded charges from failed attempts. When consumption is unknown, the confirmed subtotal is reported as a lower bound. These curves project repeated use from measured calls. Clef inference incurs no API charge. Generation and judging are accounted for separately, and local computing resources remain unpriced.

\subsection{Answer evaluation}

We evaluate answers on a fixed 200-question sample: 100 answerable and 40 unanswerable QASPER questions, plus 60 book questions. We use \method{gpt-4.1-mini-2025-04-14} to generate answers for all 11 retrieval configurations, with temperature zero and a 512-token output limit. Two controls provide either no evidence, explicitly allowing prior knowledge, or the reference evidence.

Retrieved passages receive citation labels, and overlapping source text is deduplicated. Whole passage contributions are packed in rank order within a 2,500-token content limit, including separators. Contributions that do not fit are omitted, so the generation context may contain fewer than five passages. We measure this packing loss separately. Grounded prompts instruct the generator to abstain when the supplied evidence is insufficient. Reference-evidence controls that exceed the limit are marked unavailable.

We evaluate answers with RAGAS 0.4.3 Faithfulness and Answer Accuracy, using \method{gpt-6.1-sol} as judge without revealing method names~\cite{ragas,ragasaccuracy}. Faithfulness measures support for answer claims in the actual generation context; abstentions, empty answers, and no-evidence controls are excluded. Answer Accuracy measures agreement with the fixed reference through two normalized rubric judgments, both of which must complete. It includes abstaining answers to answerable questions; unanswerable cases are assessed through abstention.

Missing answers, failed scores, and inapplicable metrics remain separate from numerical scores. Answer Accuracy is a rubric score, and neither metric establishes the correctness of individual citation labels. Development checks assess metric behavior on controlled examples; the answer evaluation remains automated, without independent human assessment.

\section{Results}
\label{sec:results}
\label{sec:decisions}
\label{sec:clef}

Across the 11 methods, the clearest differences appear on books.
Hybrid + Jev achieves the highest observed evidence coverage, while
Hybrid + Decisions obtains the highest observed Answer Accuracy among
retrievers. Full scanning does not consistently improve either result
on books and takes substantially longer.

\subsection{Evidence coverage and reduction}

Table~\ref{tab:study-retrieval} reports how much reference evidence
reaches the final five passages. On QASPER, Jev, Decisions, and Clef
all exceed 96\% coverage in both hybrid and full-scan configurations,
compared with 90.4\% for OpenAI dense retrieval. Both Clef variants
reach 98.8\%. However, the hybrid candidate pool already contains the
complete article in 395 of 396 questions, including 319 of the 320
answerable cases. QASPER therefore mainly tests passage selection;
it offers little separation between preselection and full scanning.

\begin{table}[htbp]
\centering
\small
\begin{tabular}{@{}lrr@{}}
\toprule
Method & QASPER EC@5 (\%) & Books EC@5 (\%) \\
\midrule
BM25
    & 84.7 [80.8, 88.4] & 72.1 [63.6, 80.2] \\
OpenAI dense
    & 90.4 [87.0, 93.5] & 81.3 [72.8, 89.4] \\
BGE-M3 dense
    & 89.9 [86.5, 93.0] & 74.2 [62.1, 85.6] \\
Hybrid RRF
    & 88.5 [85.0, 91.7] & 77.4 [68.6, 86.0] \\
\midrule
Hybrid + BGE
    & 88.9 [85.8, 91.9] & 86.6 [80.0, 93.1] \\
Hybrid + Jev
    & 97.0 [95.3, 98.4] & \textbf{94.1} [91.0, 97.2] \\
Hybrid + Decisions
    & 96.2 [94.1, 98.0] & 87.4 [83.4, 91.8] \\
Hybrid + Clef
    & \textbf{98.8} [97.6, 99.7] & 91.2 [84.8, 96.9] \\
\midrule
Jev full scan
    & 97.7 [96.3, 98.9] & 88.7 [81.2, 94.8] \\
Decisions full scan
    & 96.2 [94.1, 98.0] & 76.6 [70.8, 80.7] \\
Clef full scan
    & \textbf{98.8} [97.6, 99.7] & 91.6 [86.6, 96.4] \\
\bottomrule
\end{tabular}
\caption{Evidence coverage in the final five passages, averaged
within documents and then across documents. QASPER includes 320
answerable questions from 230 articles; books include 80 questions
from eight textbooks. Failed retrievals receive zero coverage.
Brackets show pointwise 95\% document-bootstrap intervals.
Bold marks the highest observed mean in each corpus.}
\label{tab:study-retrieval}
\label{tab:decisions-retrieval}
\label{tab:clef-retrieval}
\end{table}

Books expose a different tradeoff. Hybrid + Jev reaches 94.1\%
coverage, compared with 88.7\% for Jev full scan, 76.6\% for Decisions
full scan, and 91.6\% for Clef full scan. Failures contribute to these
differences: Jev and Decisions fail on four and eight book scans,
respectively, while Clef completes all 80. Among the 76 book questions
completed by both Jev variants, hybrid coverage is 94.9\% and
full-scan coverage is 93.2\%.

Full scans can still improve on dense retrieval. On books, the paired
coverage gains over OpenAI dense are 7.4 percentage points for Jev
(95\% interval: [2.0, 12.8]) and 10.3 points for Clef ([5.6, 14.8]).
Yet Clef's advantage over Jev full scan is uncertain: 2.9 points
[$-0.6$, 8.1]. The results support gains over dense retrieval without
establishing one full-scan method as best across comparisons.

\begin{figure}[!htbp]
\centering
\includegraphics[width=.96\linewidth]{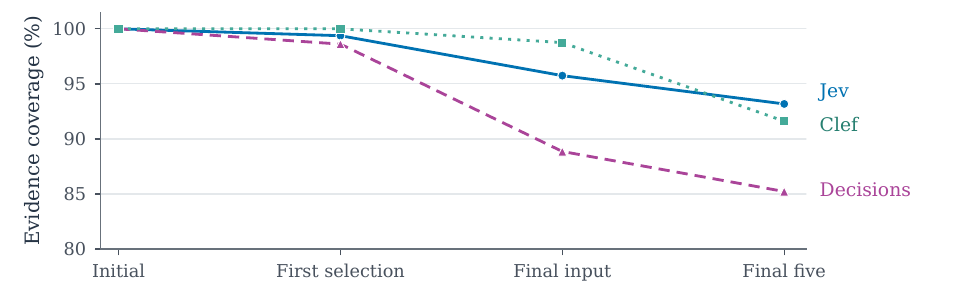}
\caption{Evidence retained through completed textbook full scans: Jev 76, Decisions 72, and Clef 80. Curves are document-macro means over each method's completed cases, not a common successful subset. Failed scans are excluded here but included in Table~\ref{tab:study-retrieval}. Initial access to every passage does not prevent loss during reduction.}
\label{fig:scan-retention}
\end{figure}

The reduction traces (Figure~\ref{fig:scan-retention}) explain why reading every passage is
insufficient. Across 76 completed Jev book scans, coverage falls from
100\% initially to 99.4\% after the first selection, 95.7\% before the
final comparison, and 93.2\% in the final five. Most loss occurs in
intermediate reductions. Clef retains 98.8\% before the final
comparison but ends at 91.6\%, placing most loss in the last selection.
Decisions ends at 85.2\% across its 72 completed scans. These traces
describe completed scans; Table~\ref{tab:study-retrieval} also includes
failed attempts.

\subsection{Latency and API cost}

On books, hybrid reranking is much faster than full scanning for all
three decision models (Table~\ref{tab:results-latency}). Median
reranking takes 1.72\,s with Jev, 4.51\,s with Decisions, and 54.69\,s
with local Clef. Full scans take 25.10\,s, 63.36\,s, and 23.3\,min,
respectively. On QASPER, hybrid and full-scan times are nearly
identical within each model, consistent with most articles fitting
in the hybrid candidate pool.

\begin{table}[htbp]
\centering
\small
\begingroup
\setlength{\tabcolsep}{4pt}
\begin{tabular}{@{}lrrrr@{}}
\toprule
& \multicolumn{2}{c}{QASPER}
& \multicolumn{2}{c}{Books} \\
\cmidrule(lr){2-3}\cmidrule(l){4-5}
Method
& Median / p95 (s) & Completed
& Median / p95 (s) & Completed \\
\midrule
Hybrid + Jev
& 0.338 / 0.751 & 396/396
& 1.724 / 2.372 & 79/80 \\
Hybrid + Decisions
& 0.621 / 1.163 & 395/396
& 4.514 / 5.577 & 80/80 \\
Hybrid + Clef
& 5.453 / 10.133 & 396/396
& 54.691 / 62.580 & 80/80 \\
\midrule
Jev full scan
& 0.339 / 0.717 & 396/396
& 25.103 / 45.711 & 76/80 \\
Decisions full scan
& 0.621 / 1.038 & 396/396
& 63.361 / 112.586 & 72/80 \\
Clef full scan
& 5.462 / 10.109 & 396/396
& 1397.672 / 2494.531 & 80/80 \\
\bottomrule
\end{tabular}
\endgroup
\caption{Median and 95th-percentile latency using the first execution
of each assigned query, with failures excluded. Completed counts are
relative to assigned queries. Hybrid times cover reranking only,
excluding candidate generation; full-scan times include selection
and reduction. Jev and Decisions allow two concurrent API calls.
Clef runs serially in BF16 on a shared GB10 and includes local
logging. Measurements describe these deployments, with different
dates and hardware conditions.}
\label{tab:results-latency}
\end{table}

The repeated end-to-end benchmark provides timing context for the
retrieval baselines. On its 50 book questions with three assigned
runs each, BM25, BGE-M3 dense, and Hybrid RRF have median retrieval
times below 0.05\,s; OpenAI dense and Hybrid + BGE take 0.56\,s and
2.05\,s. Document preparation is separate: the shared local indexes
take a mean of 56.66\,s per book, and OpenAI document embeddings take
28.98\,s. This work is reused across queries.

\begin{figure}[!htbp]
\centering
\includegraphics[width=.96\linewidth]{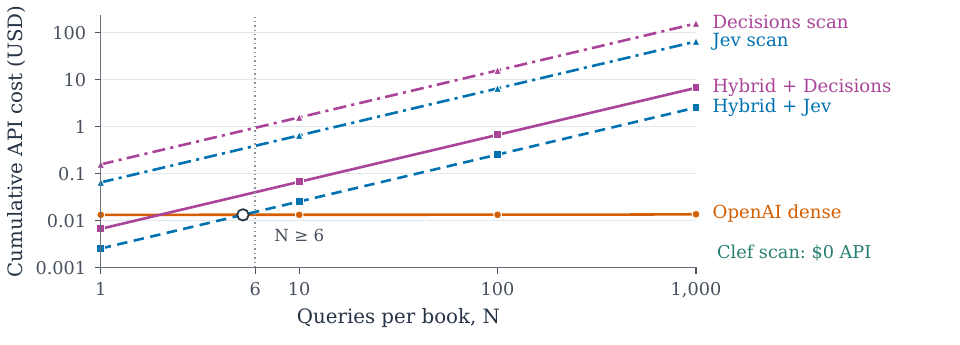}
\caption{Projected cumulative retrieval API cost per book, $C(N)=P+N\overline{Q}$, from measured preparation and query costs. Broken lines are known-charge lower bounds. Dense and Hybrid + Jev cross at $N\approx5.21$, so dense is cheaper by the sixth query; unknown charges could move the crossover earlier. Both axes are logarithmic. Clef's zero API charges are noted separately; generation, judging, and unpriced local resources are excluded.}
\label{fig:retrieval-reuse}
\end{figure}

Reuse strongly favors dense retrieval in API spending
(Figure~\ref{fig:retrieval-reuse}; Table~\ref{tab:results-cost}). For 1,000 queries per book, OpenAI
dense projects to US\$0.01351 including preparation. Hybrid + Jev
costs at least US\$2.52 and Jev full scan at least US\$64.43; the
corresponding Decisions projections are US\$6.63 and at least
US\$155.85. Under the measured book-query mix, dense retrieval
becomes cheaper than Hybrid + Jev by the sixth query per document,
even against the hybrid's lower bound based on confirmed charges.
Unresolved charges could move this crossover earlier.

\begin{table}[htbp]
\centering
\small
\begin{tabular}{@{}lrr@{}}
\toprule
Method & QASPER: $C(1000)$ & Books: $C(1000)$ \\
\midrule
OpenAI dense
    & 0.000296 & 0.01351 \\
Hybrid + Jev
    & 0.30214 & $\geq 2.52$ \\
Hybrid + Decisions
    & $\geq 0.69345$ & 6.63 \\
Jev full scan
    & 0.30537 & $\geq 64.43$ \\
Decisions full scan
    & 0.70142 & $\geq 155.85$ \\
Clef (both variants)
    & 0 & 0 \\
\bottomrule
\end{tabular}
\caption{Projected retrieval API cost in US dollars for 1,000 queries
per document, including preparation:
$C(1000)=P+1000\overline{Q}$.
Query costs average all assigned base queries within each document
and then across documents. Unknown consumption produces lower
bounds, rounded down. Clef and the other local retrieval methods
incur no API charge; local computing resources are unpriced.
Generation and judging are excluded.}
\label{tab:results-cost}
\end{table}

\subsection{Answer quality}

\begin{figure}[!htbp]
\centering
\includegraphics[width=.96\linewidth]{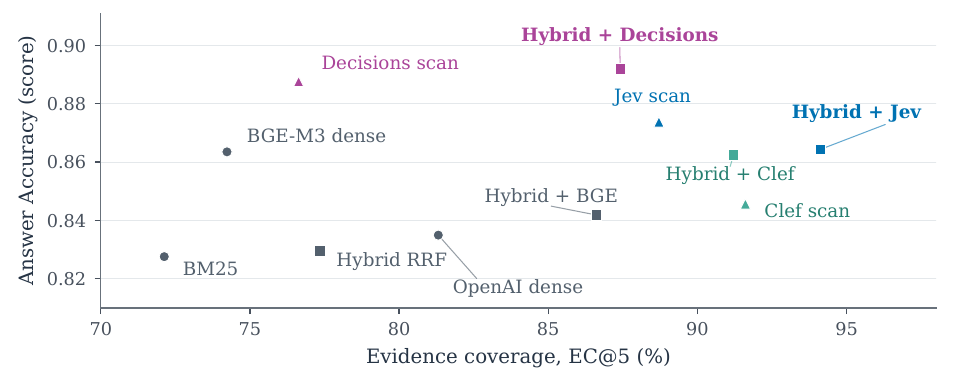}
\caption{Textbook evidence coverage versus automatic Answer Accuracy. EC@5 uses all 80 retrieval assignments; Answer Accuracy uses 53--60 successfully scored responses from 60 generation assignments per method. Both axes show document-macro means over different eligible subsets. Points are descriptive, not paired question-level observations; uncertainty in method differences is assessed by the paired comparisons in Table~\ref{tab:appendix-answer-pairs}.}
\label{fig:evidence-answer-quality}
\end{figure}

More reference evidence does not consistently produce better answers
(Figure~\ref{fig:evidence-answer-quality}).
Table~\ref{tab:study-ragas} reports automated scores for the common
answer sample. On books, Hybrid + Decisions has the highest observed
Answer Accuracy among retrievers (0.892), followed by Decisions full
scan (0.888) and Jev full scan (0.874). Hybrid + Jev leads evidence
coverage but scores 0.864 on Answer Accuracy.

\begin{table}[htbp]
\centering
\small
\begingroup
\setlength{\tabcolsep}{4pt}
\begin{tabular}{@{}lrrr@{}}
\toprule
Method & QASPER AA & Books AA & Books FA \\
\midrule
BM25
& 0.581 [100/100] & 0.828 [60/60] & 0.891 [59/59] \\
OpenAI dense
& 0.614 [99/100] & 0.835 [60/60] & 0.910 [60/60] \\
BGE-M3 dense
& 0.623 [100/100] & 0.864 [60/60] & 0.901 [60/60] \\
Hybrid RRF
& 0.592 [100/100] & 0.829 [60/60] & 0.938 [59/59] \\
\midrule
Hybrid + BGE
& 0.608 [100/100] & 0.842 [59/59] & 0.947 [59/59] \\
Hybrid + Jev
& 0.607 [99/99] & 0.864 [60/60] & 0.963 [60/60] \\
Hybrid + Decisions
& 0.609 [99/99] & \textbf{0.892} [60/60] & 0.974 [60/60] \\
Hybrid + Clef
& 0.613 [100/100] & 0.862 [59/60] & 0.962 [59/60] \\
\midrule
Jev full scan
& 0.602 [100/100] & 0.874 [57/57] & \textbf{0.975} [57/57] \\
Decisions full scan
& 0.615 [100/100] & 0.888 [53/53] & 0.973 [53/53] \\
Clef full scan
& \textbf{0.633} [100/100] & 0.846 [60/60] & 0.963 [59/60] \\
\midrule
Reference context
& 0.672 [99/100] & 0.938 [60/60] & 0.906 [60/60] \\
No evidence
& 0.052 [100/100] & 0.808 [59/60] & N/A \\
\bottomrule
\end{tabular}
\endgroup
\caption{Document-macro automated answer scores on 100 answerable
QASPER questions and 60 book questions assigned per method.
AA: Answer Accuracy; FA: Faithfulness.
Brackets show scored/requested evaluations.
Missing answers and failed evaluations are excluded from these
means; FA also requires supplied evidence and a non-abstaining
answer. Bold marks the highest observed mean among retrievers,
excluding controls.}
\label{tab:study-ragas}
\label{tab:decisions-answers}
\label{tab:clef-answers}
\end{table}

Paired comparisons support some book improvements over dense
retrieval. Hybrid + Decisions gains 0.057 in Answer Accuracy
(95\% interval: [0.019, 0.103]; 60 jointly scored questions), while
Jev full scan gains 0.048 ([0.003, 0.098]; 57 questions).
Hybrid + Decisions does not clearly exceed Hybrid + Jev: their
difference is 0.028 [$-0.005$, 0.062]. Clef full scan improves
evidence coverage over dense retrieval, but its Answer Accuracy
difference is only 0.011 [$-0.013$, 0.035]. These intervals use common
scored questions and eight book clusters; they are pointwise
comparisons without adjustment for multiple testing.

The large coverage gains on QASPER are not accompanied by a clear
improvement in answer scores. Jev full scan differs from dense
retrieval by $-0.011$ in paired Answer Accuracy [$-0.051$, 0.030],
and Clef full scan by 0.021 [$-0.018$, 0.062]. The two Clef variants
return identical ordered passages on every QASPER query, yet their
answer scores differ. That difference cannot be attributed to
retrieval: generation and judging use separate calls.

Two checks help interpret the answer scores. First, the no-evidence
control scores 0.808 on books but only 0.052 on QASPER, showing that
a high book score alone does not establish a strong dependence on
retrieved evidence. Second, context packing can remove evidence
after retrieval: across 57 completed Jev full-scan book answers,
coverage falls from 92.0\% in the retrieved passages to 90.4\%
in the supplied context.

Abstention remains limited. On the 40 unanswerable QASPER questions,
the six Jev, Decisions, and Clef configurations abstain in only
11--14 cases, compared with 17 for OpenAI dense and all 40 for the
reference-context control. High coverage on answerable questions
therefore does not establish reliable handling of questions without
supporting evidence.

\section{Limitations}
\label{sec:limitations}

The findings apply to retrieval within a known document, with up to five
passages and a 2,500-token answer context. The book evaluation covers eight
textbooks and relies largely on model-generated references and automated
answer scores. Most QASPER articles fit within the hybrid candidate pool,
limiting what this dataset reveals about the benefit of full scanning.
Coverage includes failed retrievals, while answer scores use only completed
evaluations.

Latency reflects the tested deployments: hardware, precision, concurrency,
and hosted service conditions differ across methods. Hosted model behavior
can also vary between calls. API charges exclude local computing resources,
including Clef's GPU use, so they do not represent total operating cost.

\section{Conclusion}

We compared 11 retrieval configurations on research articles and complete
textbooks. On books, Hybrid + Jev achieved the highest observed evidence
coverage (94.1\%), while Hybrid + Decisions obtained the highest observed
Answer Accuracy among retrievers (0.892). Both Clef variants reached
98.8\% evidence coverage on QASPER. Across the comparisons, recovering more
reference evidence did not consistently improve answer quality.

Full scanning reached every passage initially but lost evidence during
reduction. On books, hybrid reranking was much faster and achieved the
highest observed coverage and answer scores. For 1,000 queries per document,
dense retrieval had much lower projected spending than the paid decision
APIs, including its initial embedding cost. Choosing a retrieval method
therefore requires considering evidence retention, answer quality, and
cost over repeated use together.

\clearpage
\appendix

\section{Corpus and reproducibility}
\label{sec:appendix-data}
\label{sec:book-cohort}

Table~\ref{tab:corpus-details} lists the complete textbooks and the
subsets used for timing repetitions and answer evaluation. Each book has
ten questions: four factual, three explanatory, and three requiring
multiple sections. The eight test books contain 4.46 million tokens and
10,508 passages. Development uses two separate books. The test subsets
are fixed across methods, with 11 retrieval configurations and two answer
controls.

\begin{table}[htbp]
\centering\small
\begin{tabular}{@{}lrrrr@{}}
\toprule
Book & Tokens & Passages & Timing Q. & Answer Q. \\
\midrule
\multicolumn{5}{@{}l}{\textit{Test: 80 questions}} \\
Astronomy 2e
    & 749,826 & 1,765 & 4 & 9 \\
Biology 2e
    & 954,082 & 2,245 & 4 & 9 \\
Psychology 2e
    & 515,399 & 1,213 & 9 & 6 \\
Principles of Economics 3e
    & 635,781 & 1,496 & 8 & 8 \\
American Government 3e
    & 509,472 & 1,199 & 8 & 5 \\
Introduction to Business
    & 420,034 & 989 & 8 & 8 \\
Introduction to Sociology 3e
    & 398,802 & 939 & 4 & 8 \\
Introduction to Philosophy
    & 281,392 & 662 & 5 & 7 \\
\midrule
Total test
    & 4,464,788 & 10,508 & 50 & 60 \\
\midrule
\multicolumn{5}{@{}l}{\textit{Development: 20 questions}} \\
U.S. History
    & 605,929 & 1,426 & --- & --- \\
Introduction to Anthropology
    & 454,904 & 1,071 & --- & --- \\
\bottomrule
\end{tabular}
\caption{Book sizes and test subsets. Timing questions receive three
assigned retrieval executions; this column does not define the full-test
primary timing comparison used for Jev, Decisions, and Clef. Answer
questions are shared by all 13 answer arms. Development books are outside
these test subsets. Token counts use the common text tokenizer.}
\label{tab:corpus-details}
\end{table}

The QASPER test contains 396 questions from 264 articles. Evidence
coverage uses its 320 answerable questions from 230 articles; costs and
execution counts include all 396 questions. The exact oracle reaches
100\% reference coverage with at most five passages for every answerable
test question: 320 from QASPER and 80 from books. Thus, the five-passage
limit permits complete reference recovery in this dataset. Fitting those
passages into the answer context remains a separate constraint.

The recorded checkpoint prefixes are \texttt{5617a9f61b02} for BGE-M3,
\texttt{953dc6f6f85a} for the BGE reranker, and \texttt{ed3eed331870} for
Clef. Clef uses SDPA attention and the PyTorch CUDA convolution fallback,
without quantization or parameter offloading to the CPU. Full revision
identifiers, environments, prompts, source locks, and execution records
are documented in the \href{https://github.com/jrterven/jev-vs-rag}{code repository}.

The evaluation code was checked with synthetic examples covering
paraphrases, contradictions, missing support, partial answers, and
abstention. All eight Faithfulness groups and eleven Answer Accuracy
groups met their expected ranges in three repetitions. A separate
Factual Correctness diagnostic failed one condition and was not used for
test scoring. A separate model also reviewed the 100 development and test
book questions, reference answers, and 148 supporting spans without
seeing retrieval outputs or answer scores. It recorded three minor
observations and no confirmed reference error; no questions or evidence
spans were changed. These checks document expected metric behavior and
reference consistency.

\section{Retrieval diagnostics and resource use}
\label{sec:appendix-retrieval}
\label{sec:primary-additional}

All 5,236 primary retrieval assignments have a recorded outcome: 5,222
completed and 14 failed. Table~\ref{tab:study-retrieval-accounting}
identifies the failures and the valid repeated timing observations.
Failed primary retrievals remain in the evidence evaluation with zero
coverage. A failed or invalid timing observation is excluded from latency
statistics.

\begin{table}[htbp]
\centering\small
\begingroup
\setlength{\tabcolsep}{5pt}
\begin{tabular}{@{}lrrr@{}}
\toprule
Method
& \shortstack{Primary failures\\QASPER / books}
& \shortstack{Valid repeated runs\\QASPER / 150}
& \shortstack{Valid repeated runs\\Books / 150} \\
\midrule
BM25               & 0 / 0 & 150 & 150 \\
OpenAI dense       & 0 / 0 & 150 & 150 \\
BGE-M3 dense       & 0 / 0 & 150 & 150 \\
Hybrid RRF         & 0 / 0 & 150 & 149 \\
\midrule
Hybrid + BGE       & 0 / 0 & 150 & 150 \\
Hybrid + Jev       & 0 / 1 & 143 & 146 \\
Hybrid + Decisions & 1 / 0 & 150 & 150 \\
Hybrid + Clef      & 0 / 0 & --- & --- \\
\midrule
Jev full scan       & 0 / 4 & 143 & 144 \\
Decisions full scan & 0 / 8 & 150 & 138 \\
Clef full scan      & 0 / 0 & --- & --- \\
\bottomrule
\end{tabular}
\endgroup
\caption{Primary failures and repeated end-to-end timing availability.
Every method has 396 primary QASPER and 80 primary book assignments.
The repeated benchmark assigns 50 questions per corpus and three runs
per question. Hybrid RRF has one completed run with invalid timing;
other exclusions in the repeated-run columns are failed runs. Dashes
indicate that Clef has no comparable completed repetition summary.}
\label{tab:study-retrieval-accounting}
\end{table}

The Decisions repetition counts include valid runs under both exclusive
and shared GPU conditions; their latency summaries remain separate by
condition. Clef completed all 952 primary searches, but only 18 of 400
additional timing repetitions, with 382 incomplete. Its reported latency
therefore uses primary executions only. These omissions do not count as
primary retrieval failures. Service interruptions and varying deployment
conditions remain part of the timing records.

The shared hybrid pool of up to 50 candidates already contains 98.75\%
of book reference evidence before reranking, averaged across all 80
questions and eight books. This coverage refers to the entire candidate
pool, without the final five-passage restriction.
Table~\ref{tab:appendix-retention} shows where full scans lose evidence.
Jev and Decisions lose most during intermediate reductions; Clef loses
most when selecting the final five. These traces include completed scans
only; primary means also include failed retrievals.

\begin{table}[htbp]
\centering\small
\begin{tabular}{@{}lrrrrr@{}}
\toprule
Full scan & Completed & Initial & First selection & Final input & Final five \\
\midrule
Jev       & 76/80 & 100.00 & 99.38  & 95.75 & 93.18 \\
Decisions & 72/80 & 100.00 & 98.61  & 88.86 & 85.23 \\
Clef      & 80/80 & 100.00 & 100.00 & 98.75 & 91.61 \\
\bottomrule
\end{tabular}
\caption{Book evidence coverage (\%) through the reduction stages,
averaged within documents and then across eight books. Each row uses
that method's completed questions; the rows are not restricted to a
common successful subset. Stages describe candidate sets, and no
confidence intervals are inferred for these trace summaries.}
\label{tab:appendix-retention}
\end{table}

Leaving out one book at a time gives a Jev full-scan advantage over
OpenAI dense of 5.8--8.9 percentage points. Against Hybrid + BGE, the
range is 0.2--4.3 points, but the full-sample 95\% bootstrap interval
[$-5.6$, 9.7] includes zero. These omission ranges describe variation
across book subsets and are distinct from the paired bootstrap intervals.

Preparation is a one-time document cost
(Table~\ref{tab:study-preparation}). Local preparation covers the shared
lexical and BGE indexes, including initial model loading. It is not
multiplied by the number of methods that reuse those indexes.

\begin{table}[htbp]
\centering\small
\begin{tabular}{@{}lrrr@{}}
\toprule
Corpus & Documents & Local mean / median (s) & OpenAI mean / median (s) \\
\midrule
QASPER & 264 & 0.60 / 0.47   & 1.14 / 0.64 \\
Books  & 8   & 56.66 / 51.49 & 28.98 / 23.68 \\
\bottomrule
\end{tabular}
\caption{Document preparation, measured separately from retrieval and
answer generation. Local time covers shared indexes; OpenAI time covers
document embedding calls.}
\label{tab:study-preparation}
\end{table}

Table~\ref{tab:appendix-query-cost} gives the per-query inputs to the
reuse projection. Only OpenAI dense has a paid preparation term:
US\$0.000109950985 per article and US\$0.0131306925 per book.
Other methods may require local preparation, which is not priced here.
Charges from failed base queries are included when known; unknown
consumption produces a lower bound. Timing repetitions, answer
generation, and judging are outside this projection.

\begin{table}[htbp]
\centering\small
\begin{tabular}{@{}lrr@{}}
\toprule
Method & QASPER: USD/query & Books: USD/query \\
\midrule
OpenAI dense
    & $1.8625\times10^{-7}$ & $3.8050\times10^{-7}$ \\
Hybrid + Jev
    & 0.00030214 & $\geq 0.00252247$ \\
Hybrid + Decisions
    & $\geq 0.00069345$ & 0.00662575 \\
Jev full scan
    & 0.00030537 & $\geq 0.06443530$ \\
Decisions full scan
    & 0.00070142 & $\geq 0.15585864$ \\
Clef (both variants)
    & 0 & 0 \\
\bottomrule
\end{tabular}
\caption{Mean retrieval API cost per assigned base query, averaging
within documents and then across documents. Values use the rates
recorded during execution; lower bounds are rounded down. The other
local retrieval methods also have zero API charges, with computing
resources unpriced.}
\label{tab:appendix-query-cost}
\end{table}

For books, the crossover between dense retrieval and the Hybrid + Jev
lower bound based on confirmed charges is
\[
\frac{0.0131306925}
{0.002522472225-0.0000003805}
=5.2063\ \text{queries}.
\]
Thus, dense retrieval is cheaper by the sixth query per book under this
projection. Unknown Jev charges can move the crossover earlier.

\section{Answer comparisons and evaluation outcomes}
\label{sec:appendix-answers}
\label{sec:final-judge}

Table~\ref{tab:appendix-answer-pairs} collects paired Answer Accuracy
comparisons against dense retrieval and between decision models.
Each comparison uses only questions scored for both methods and then
averages within documents. Consequently, these effects cannot be
reconstructed by subtracting the conditional means in the main results.
Book comparisons use eight documents. Intervals are descriptive and
are not adjusted for multiple comparisons.

\begin{table}[htbp]
\centering\small
\begingroup
\setlength{\tabcolsep}{4pt}
\renewcommand{\arraystretch}{1.15}
\begin{tabular}{@{}p{.31\linewidth}rr@{}}
\toprule
Comparison
& QASPER: difference [95\% CI]; $n$
& Books: difference [95\% CI]; $n$ \\
\midrule
Jev scan $-$ OpenAI dense
& $-0.011$ [$-0.051$, 0.030]; 99
& $+0.048$ [0.003, 0.098]; 57 \\
Hybrid + Decisions $-$ OpenAI dense
& $-0.002$ [$-0.041$, 0.038]; 98
& $+0.057$ [0.019, 0.103]; 60 \\
Decisions scan $-$ OpenAI dense
& $+0.003$ [$-0.036$, 0.043]; 99
& $+0.043$ [0.015, 0.075]; 53 \\
Hybrid + Clef $-$ OpenAI dense
& $+0.000$ [$-0.038$, 0.039]; 99
& $+0.028$ [$-0.009$, 0.065]; 59 \\
Clef scan $-$ OpenAI dense
& $+0.021$ [$-0.018$, 0.062]; 99
& $+0.011$ [$-0.013$, 0.035]; 60 \\
\midrule
Hybrid + Decisions $-$ Hybrid + Jev
& $+0.001$ [$-0.028$, 0.031]; 98
& $+0.028$ [$-0.005$, 0.062]; 60 \\
Decisions scan $-$ Jev scan
& $+0.014$ [$-0.012$, 0.040]; 100
& $+0.013$ [0.004, 0.026]; 50 \\
Hybrid + Clef $-$ Hybrid + Decisions
& $+0.003$ [$-0.027$, 0.034]; 99
& $-0.032$ [$-0.046$, $-0.017$]; 59 \\
Clef scan $-$ Decisions scan
& $+0.018$ [$-0.009$, 0.046]; 100
& $-0.038$ [$-0.064$, $-0.014$]; 53 \\
\bottomrule
\end{tabular}
\endgroup
\caption{Paired Answer Accuracy differences, first method minus second.
``Scan'' denotes full scan. Each cell reports the difference, its 95\%
interval, and the number of jointly scored questions $n$. Intervals use
10,000 whole-document bootstrap resamples with seed 42. An interval
containing zero does not establish equivalence.}
\label{tab:appendix-answer-pairs}
\label{tab:decisions-answer-pairs}
\label{tab:clef-answer-pairs}
\end{table}

The paired book Faithfulness gains for Jev full scan are 0.069
[0.050, 0.090] over OpenAI dense (57 pairs), and 0.028 [0.010, 0.050]
over Hybrid + BGE (56 pairs). These comparisons apply to answers with
supplied evidence that did not abstain. They do not include every
assigned answer.

Across the 13 answer arms, 2,586 of 2,600 assignments produced usable
answers. Ten missing answers followed retrieval failures: three for Jev
full scan and seven for Decisions full scan, all on books. Four others
resulted from generation errors, one each for BGE-M3 on unanswerable
QASPER, Hybrid + BGE on books, and Hybrid + Jev and Hybrid + Decisions on
answerable QASPER. All reference-context controls were eligible.
Table~\ref{tab:study-answer-outcomes} distinguishes completed answers from
abstentions in each group.

\begin{table}[htbp]
\centering\small
\begin{tabular}{@{}lrrr@{}}
\toprule
Method
& \shortstack{QASPER answerable\\100 assigned}
& \shortstack{Books\\60 assigned}
& \shortstack{QASPER unanswerable\\40 assigned} \\
\midrule
BM25               & 100 (2) & 60 (1) & 40 (12) \\
OpenAI dense       & 100 (1) & 60 (0) & 40 (17) \\
BGE-M3 dense       & 100 (2) & 60 (0) & 39 (16) \\
Hybrid RRF         & 100 (2) & 60 (1) & 40 (12) \\
\midrule
Hybrid + BGE       & 100 (1) & 59 (0) & 40 (15) \\
Hybrid + Jev       & 99 (1)  & 60 (0) & 40 (12) \\
Hybrid + Decisions & 99 (1)  & 60 (0) & 40 (12) \\
Hybrid + Clef      & 100 (1) & 60 (0) & 40 (12) \\
\midrule
Jev full scan       & 100 (1) & 57 (0) & 40 (13) \\
Decisions full scan & 100 (1) & 53 (0) & 40 (14) \\
Clef full scan      & 100 (2) & 60 (0) & 40 (11) \\
\midrule
Reference context  & 100 (6)  & 60 (0) & 40 (40) \\
No evidence        & 100 (50) & 60 (0) & 40 (18) \\
\bottomrule
\end{tabular}
\caption{Each cell gives completed answers, with abstentions in
parentheses. Abstention is an observed response behavior; these are
question counts, not document-macro percentages. Rates use completed
answers as the denominator: BGE-M3, for example, abstains in 16 of its
39 completed unanswerable cases. Missing answers remain separate.}
\label{tab:study-answer-outcomes}
\end{table}

The judge scored 4,233 of 4,244 requested evaluations; 11 failed.
Table~\ref{tab:study-all-metric-accounting} shows the full accounting,
including unanswerable questions. Faithfulness is not requested for
abstentions or the no-evidence control; Answer Accuracy is not requested
for unanswerable references. The 928 inapplicable answer--metric
combinations and all missing or failed outputs remain outside numerical
scores. No requested evaluation remains pending or unscorable.

\begin{table}[htbp]
\centering\small
\begin{tabular}{@{}llrrr@{}}
\toprule
Questions & Metric & Scored / requested & N/A & Errors \\
\midrule
QASPER answerable
    & Faithfulness    & 1,172 / 1,177 & 121 & 5 \\
    & Answer Accuracy & 1,296 / 1,298 & 0   & 2 \\
Books
    & Faithfulness    & 705 / 707     & 62  & 2 \\
    & Answer Accuracy & 767 / 769     & 0   & 2 \\
QASPER unanswerable
    & Faithfulness    & 293 / 293     & 226 & 0 \\
    & Answer Accuracy & 0 / 0         & 519 & 0 \\
\bottomrule
\end{tabular}
\caption{Evaluation outcomes across all 13 answer arms. N/A counts
metrics not requested for completed answers; it excludes missing
generations. Requested evaluations equal scored evaluations plus
errors. Failed and inapplicable metrics receive no imputed score.}
\label{tab:study-all-metric-accounting}
\end{table}
\end{document}